\def\year{2022}\relax
\documentclass[letterpaper]{article} 
\usepackage{aaai22}  
\usepackage{times}  
\usepackage{helvet}  
\usepackage{courier}  
\usepackage[hyphens]{url}  
\usepackage{graphicx} 
\usepackage{natbib}  
\usepackage{caption} 
\usepackage{array,multirow}
\DeclareCaptionStyle{ruled}{labelfont=normalfont,labelsep=colon,strut=off} 
\usepackage{algorithm}
\usepackage{algorithmic}

\usepackage{newfloat}
\usepackage{listings}
\floatstyle{ruled}
\newfloat{listing}{tb}{lst}{}
\floatname{listing}{Listing}
\usepackage{color}

\usepackage{booktabs}
\usepackage{enumitem}

\title{Beyond Value: \textsc{CheckList} for Testing Inferences in Planning-Based RL}

\author {
    Kin-Ho Lam, 
     Delyar Tabatabai, 
     Jed Irvine,
     Donald Bertucci,
     Anita Ruangrotsakun,\\
     Minsuk Kahng,
     Alan Fern
}
\affiliations {
    Oregon State University\\
    \{lamki, tabatase, jed.irvine, bertuccd, ruangroc, minsuk.kahng, alan.fern\}@oregonstate.edu
}

\begin{document}

\maketitle

\begin{abstract}
Reinforcement learning (RL) agents are commonly evaluated via their expected value over a distribution of test scenarios.
Unfortunately, this evaluation approach provides limited evidence for post-deployment generalization beyond the test distribution. In this paper, we address this limitation by extending the recent \textsc{CheckList} testing methodology from natural language processing to planning-based RL. Specifically, we consider testing RL agents that make decisions via online tree search using a learned transition model and value function. The key idea is to improve the assessment of future performance via a \textsc{CheckList} approach for exploring and assessing the agent's inferences during tree search. The approach provides the user with an interface and general query-rule mechanism for identifying potential inference flaws and validating expected inference invariances. We present a user study involving knowledgeable AI researchers using the approach to evaluate an agent trained to play a complex real-time strategy game. The results show the approach is effective in allowing users to identify previously-unknown flaws in the agent's reasoning. In addition, our analysis provides insight into how AI experts use this type of testing approach, which may help improve future instantiations.
\end{abstract}

\section{Introduction}

Evaluating reinforcement learning (RL) agents is typically done by estimating a single quantity, the expected value, via Monte-Carlo simulation over a set of validation scenarios. However, this single quantity alone provides little insight into the agent's underlying behavior and reasoning. As a result, this evaluation methodology may not uncover flaws in the agent that hurt generalization to reasonable post-deployment situations. For example, an agent may learn behaviors that maximize reward by abusing details of the simulator used for training and validation \cite{baker2020emergent}, learn aberrant behaviors unrelated to the general task, or the training reward function may not always relate to task accomplishment after deployment \cite{clark_2019,gleave2021adversarial}. Further, for planning-based RL agents, a systemic error in the model may go undetected during evaluation, but still manifest in rare, but serious, post-deployment failures. 

One approach to improving confidence in RL agents during validation is to produce explanations of agent decisions that can be evaluated by a human \cite{puiutta2020,heuillet2021}.
There are a wide range of explanation mechanisms for RL, which primarily focus on explaining learned reactive policies. For example, explanations may highlight the most salient parts of an agent's input (e.g. \cite{pmlr-v80-greydanus18a,mott2019}), extract interpretable structure of the agent's policy \cite{MISCkoul2018learning,verma2018,asai2020,madumal2020}, provide a rationale for preferring one action over another \cite{Waa2018,juozapaitis2019,lin2021}, or visualize the internal representations~\cite{wang2018dqnviz,tabatabai2021did,mishra2022not}.
Unfortunately, these and most other types of explanations in RL provide no insight into the internal ``reasoning steps'' that result in action selection. This limits evaluation to the level of overall decisions/actions, for example, noticing that an agent paid attention to seemingly irrelevant information when selecting a particular action. 

One way to provide explanations at the level of reasoning steps is to consider planning-based RL agents, which plan using learned models and control knowledge. In concept, this allows for human-validation of an agent's internal reasoning used for action selection. This approach, however, raises at least two challenges.

First, such agents may perform at a superhuman level where the reasoning is too complex for human consumption (e.g. enormous minimax trees). While work on explainable planning \cite{fox2017,chakraborti2020} attempts to mitigate this issue, humans ultimately have limited capacity. In this work, we address this issue by assuming the RL agents use sound planning algorithms (e.g. minimax), which means mistakes are due to inaccuracies in the learned models and/or control knowledge. This implies humans can focus validation effort towards building confidence in the prediction accuracy of learned components, rather than understanding how the planner combines the predictions into an overall decision. 

Second, the sheer volume of information to be validated (e.g. thousands to millions of possible actions considered by RL agents) raises the question of how humans can efficiently inspect and analyze them.
We address this challenge by taking inspiration from the recent \textsc{CheckList} methodology for testing learned systems for natural language processing (NLP) \cite{ribeiro2020accuracy}. \textsc{CheckList} allows human experts or developers to create rules that capture invariants or other properties the learned system should never satisfy. In this way, rules provide an empirical and reproducible validation metric for humans to identify flaws in NLP systems.

Our main contribution is an adaptation of \textsc{CheckList} for validation of planning-based RL agents, which we refer to as \emph{\textsc{CheckList} for RL (C4RL)}. While simple in concept, we are unaware of prior work that operationalizes and evaluates such an approach in the context of RL. C4RL allows a human to use their domain and RL-architecture knowledge to create logical assertions, called \emph{query-rules}. A query-rule is a model-agnostic, domain-specific relational-algebra expression that asserts a property the agent's reasoning should never, or very rarely, satisfy. For example, query-rules might specify known invariants or pairwise action-value orderings of a learned value function. We present three general classes of query-rules that can be used for testing reasoning traces produced during validation runs as well as counterfactual situations derived from the traces.

Our second contribution is a formative user study to observe how one might use C4RL to find agent reasoning flaws. 
Participants were provided with an interface to validate a complex real-time strategy (RTS) game played by a planning-based agent, while creating and evaluating query-rules. 
They successfully formed a range of query-rules that identify known and previously-unknown flaws in the agent's reasoning.

\section{RL Architecture and Game Environment}

While the C4RL methodology is domain and model agnostic and can be instantiated for different application domains and types of RL agents, for concreteness, we present C4RL in the context of a game environment and RL architecture developed for our user study.
This section describes our real-time strategy game, Tug-of-War, and RL agent used. 

\subsection{Experimental Domain: Tug-of-War}
\label{gameRules-information}

\emph{Tug-of-War (ToW)} is an adversarial two-player RTS game, built on using the StarCraft 2 game engine. ToW is a challenging domain for RL because of the large state and action spaces, complex dynamics, sparse reward, etc. 
We used the ToW environment we developed in prior work~\cite{lam2021identifying}. Below we provide an overview.

\begin{figure}[!tb]
    \centering
    \includegraphics[width=0.9\columnwidth]{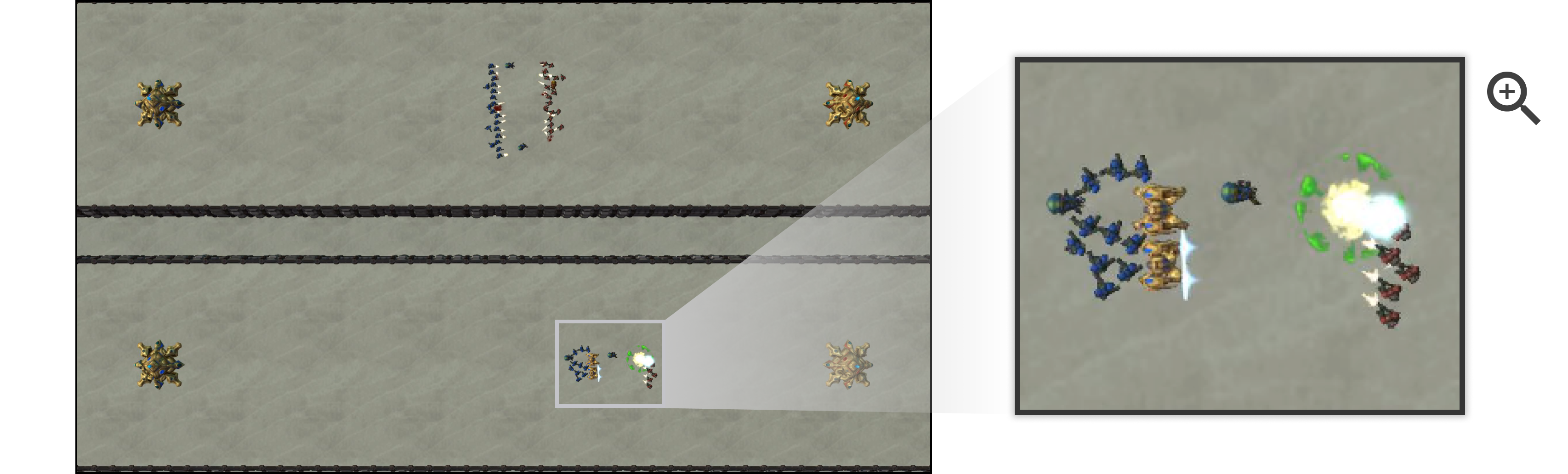}
    \vspace{-5pt}
    \caption{
    The Tug of War map is divided into two separate lanes, top and bottom.
    The Friendly AI (left) and Enemy AI (right) own bases (gold star-shaped buildings) on opposite sides of the map.
    Each round, troops from the opposing player automatically march towards their opponent's side of the map and attack the closest enemy in their lane. 
    }
    \label{fig:ToW-ScreenShot}
\end{figure}

In ToW, two players, Friendly AI (on the left) and Enemy AI (on the right), adversarily play against each other.
The map is divided into the top and bottom lanes, where player's bases are placed on opposite ends (Figure \ref{fig:ToW-ScreenShot}). The game proceeds in 30 second waves, where before each wave, each player may select either the top or bottom lane and decide which military-unit production buildings to build in the selected lane with their available currency. 

At the start of each wave, each production building produces one unit of the specified type. The units automatically walk across the lane toward the enemy base and automatically attack incoming enemy units or the opponent's base if close enough. Each unit has an initial amount of health that decreases when attacked until no health is remaining and the unit disappears. The three unit types, Marines, Immortals, and Banelings, form a rock-paper-scissors relationship with respect to the amount of damage done when one unit attacks another. The first player to destroy one of the opponent bases wins. If no base is destroyed after 40 waves (20 minutes), the player with the lowest health base loses. 

Importantly, players do not control the detailed movement and target selection of individual units as they move across the map, but rather control only the higher-level choice of how to spend resources each wave. Thus, the possible actions/choices available at each wave are the different purchase combinations that can be afforded with the current resources, which can range from 10s to 1000s of possible actions. At any moment of the game there can be 10s to 100s of units on the map, which  creates an enormous state space.

Preceding each wave, the RL agents observe the state of the map and select an action that will dictate resource spending at the start of the next round.
In concept, the observations provide perfect state information.
The game dynamics have randomness due to variations in unit movements, organization, and attacks, making future states and opponent actions difficult to predict. The sparse reward is zero for both players at each decision point until the end of the game where the winning player receives +1 reward and the losing player receives 0.
The ToW discrete state space describes the quantity of each troop on the field, the health of the bases, current buildings owned by each player, and available currency to purchase buildings.
The discrete action space describes the quantity of each building type a player may purchase.

\subsection{Planning-Based Agent Architecture}
\label{sec:architecture}

We used an agent we developed in prior work, and below we overview its architecture \cite{lam2021identifying}.
This planning-based RL architecture makes decisions via tree search using a learned game dynamics model and leaf evaluation function. We use the terminology ``planning-based RL" rather than ``model-based RL" to emphasize that decisions are made via deliberative planning. This is in contrast to many model-based RL agents, which primarily use the learned model to train a reactive policy with additional simulated experience (e.g. \cite{sutton1990, kaiser2020}). The search trees produced by the planning-based architecture can serve as a type of explanation for each decision and will be the main artifacts being validated via C4RL.

The agent architecture is similar to AlphaZero \cite{silver2018}, an RL agent based on game-tree search, except that our agent uses a learned model rather than an exact model for tree construction.  
Our agent searches over human-interpretable abstract states, each providing information about the health of bases, building counts, the number of friendly/enemy troops of each type in each of four evenly divided grid cells per lane, etc.
This abstraction (see Figure \ref{fig:UI}-C-1) is rich enough for humans to understand the states and make reasonable decisions. 

At each decision point, an action is selected by building a minimax search tree using three learned components: 1) \emph{Transition Model}, which predicts the next abstract state (i.e., 30 seconds after) given a current abstract state and both players' actions; 
2) \emph{Action Ranker}, which returns a numeric ranking over actions in an abstract state based on their estimated action values; and 3) \emph{State-Value Function}, which returns a value estimate (probability of winning) given an abstract state. 
The transition model supports building the tree starting at the current abstract state which becomes the root node. 
The action ranker prunes actions from the tree among many possible friendly and enemy action combinations, to include between 20 to 5 friendly AI's actions and between 10 to 3 enemy's actions depending on the depth of the tree.
We specified the depth of the tree as 2, meaning that the tree is a two-step look-ahead.
Figure \ref{fig:UI}-C illustrates a subset of the tree: the leftmost node is the root (C-1); six action combinations are shown next to the root (C-2); the next states are predicted by the agent (C-3); and the rightmost node is a leaf of the tree at depth 2.

\section{\textsc{CheckList} for Planning-Based RL}
\label{sec:checklist}

Our work takes inspiration from the recent \textsc{CheckList} methodology for validating natural language processing (NLP) systems.
We first overview \textsc{CheckList} for NLP, and then describe our adaptation for planning-based RL agents.

\subsection{Background: \textsc{CheckList} for NLP}

State-of-the-art in NLP is dominated by machine-learning approaches trained on large data corpuses. Validation of such NLP systems is complicated by both their black-box nature and the vastness of possible natural-language inputs. Recently, \citet{ribeiro2020accuracy} proposed \textsc{CheckList} as a methodology for validating NLP systems and demonstrated its potential. At a high level, \textsc{CheckList} directs NLP developers to use behavioral-testing approaches from software engineering to identify categories of errors (e.g. violating prediction invariance to types of input perturbations). A \emph{domain-specific language (DSL)} is used to define abstract test cases that via rules that can generate a large corpus of ground test cases for validation. This approach allows a developer or tester to quickly produce thousands of valid and reproducible tests for an NLP system using their linguistic, domain, and NLP knowledge.

In more detail, \textsc{CheckList} allows human testers to define \emph{test types} that specify classes of inputs (e.g. sentences of a specific form) and desired outputs (e.g. a sentiment classification). The test types are instantiated via human-created \emph{test templates}, that specify general sentence structures with variables for which specified sets of words can be substituted. 
For example, in sentiment analysis, a template may transform a pre-existing ``positive" sentiment test sentence to a negated form, which should result in the system producing a ``negative" sentiment classification. This approach demonstrated human testers were able to effectively bring large quantities of previously unknown failures in NLP sentiment-analysis systems to the attention of developers.

\begin{table*}[!t]
\centering 
\begin{tabular}{p{1.1in} | p{5.5in} }
\toprule
Query Class & Example Query Rule   \\
\midrule
\textbf{Static State Rules}:

Validate a single output state &
\begin{tabular}[t]{@{} p{3.75in} l @{}}
\textit{Example 1-1.}
This outputs state produced by the agent places a troop (immortal) on the field although there are no buildings to produce it.

\texttt{\small outputState.friendlyImmortalBldgsTop = 0 AND (}

\texttt{\small outputState.friendlyImmortalTopGrid1 + outputState.friendlyImmortalTopGrid2 + ... + outputState.friendlyImmortalTopGrid4) > 0}
&
\raisebox{-0.9\height}{\includegraphics[width=4cm]{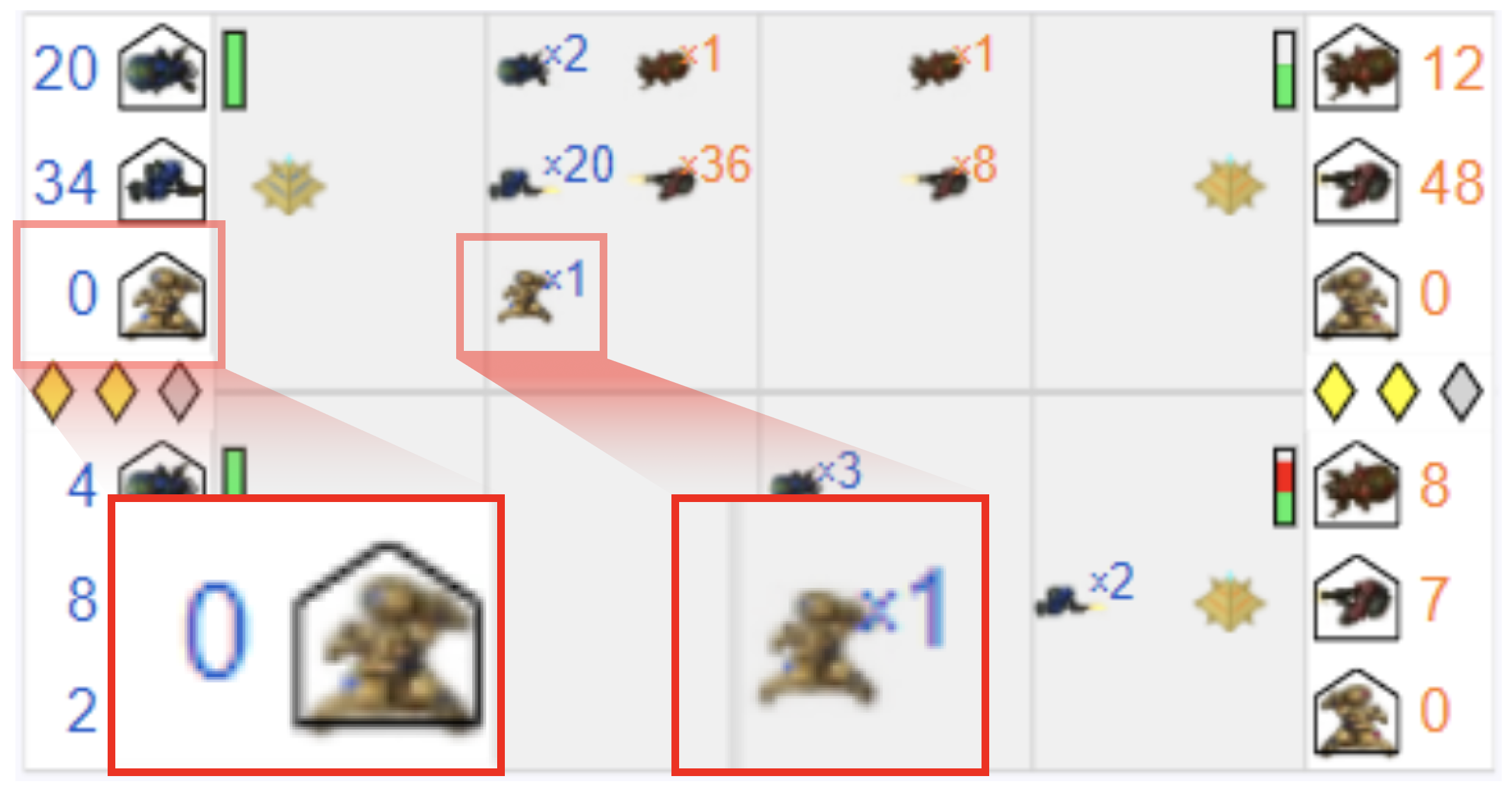}} \label{fig:immortals_no_buildings} 
\end{tabular}
\\
\cmidrule{2-2}
 &
\begin{tabular}[t]{@{} p{3.9in} l @{}}
        \textit{Example 1-2.} The agent assigns a non-zero probability of winning the game by destroying the enemy's bottom base despite having already lost. 
 \texttt{\small outputState.friendlyHealthTop = 0 AND}
 
 \texttt{\small (winProb.probabilityOfWinInTopLane +}
 
 \texttt{\small winProb.probabilityOfWinInBottomLane) != 0}  
          & 
 \raisebox{-0.8\height}{\includegraphics[trim=0 0.8in 0 0,clip,width=3.4cm]{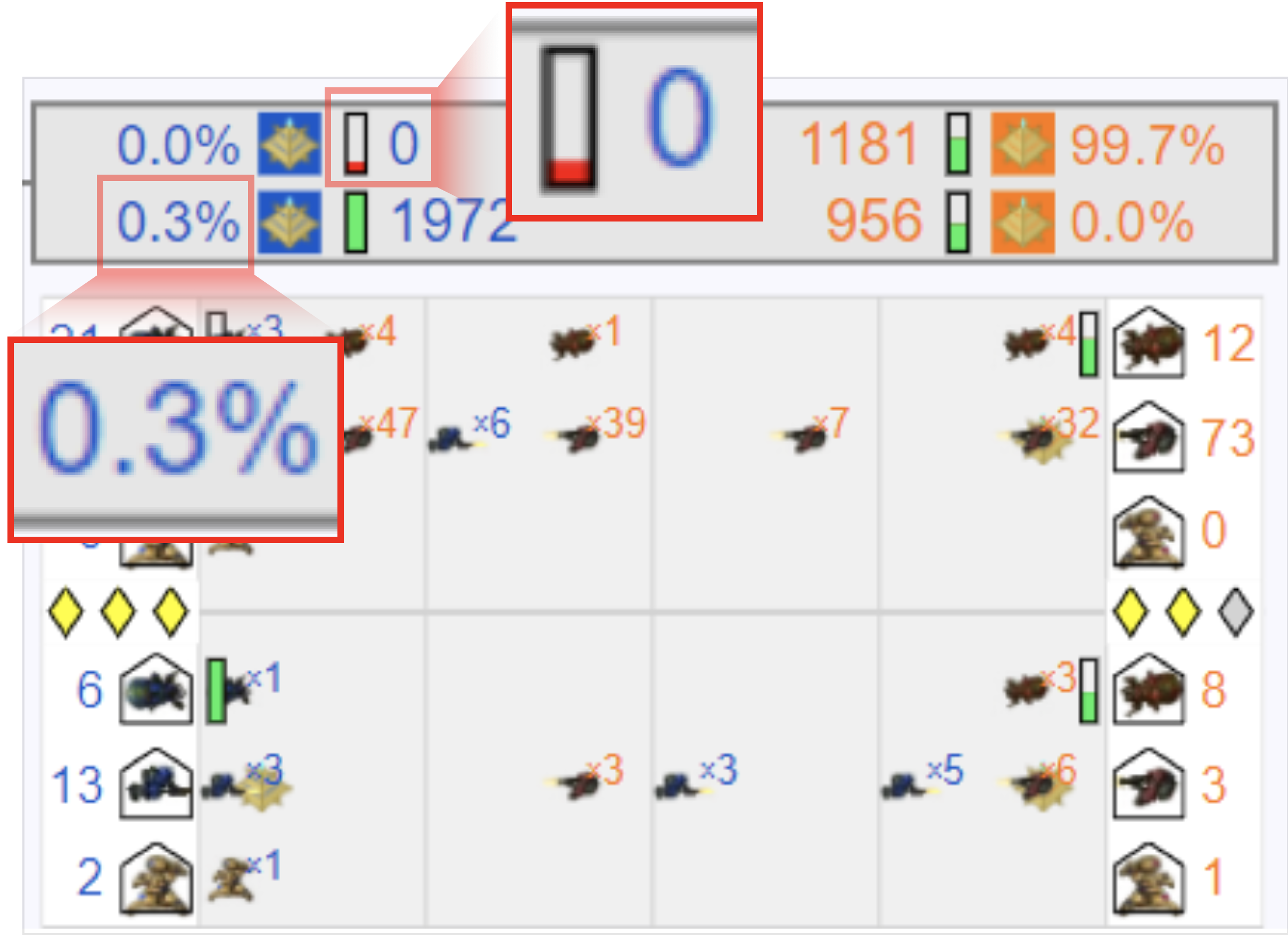}} \label{fig:nonzero_lose}
     \end{tabular}
\\
\midrule
\textbf{Transition Rules}:
Validate the relationships between an output state and its input. 
&
\begin{tabular}[t]{@{}p{3.05in} l}
\textit{Example 2-1.}
Base health cannot increase due to game rules. In this example, the output state's base health is higher than the input's.
See Section \ref{sec:caseStudy2-1} for more details.

\texttt{\small  outputState.enemyHealthTop - inputState.enemyHealthTop > 5.0}
&
\raisebox{-0.85\height}{\includegraphics[width=5.7cm]{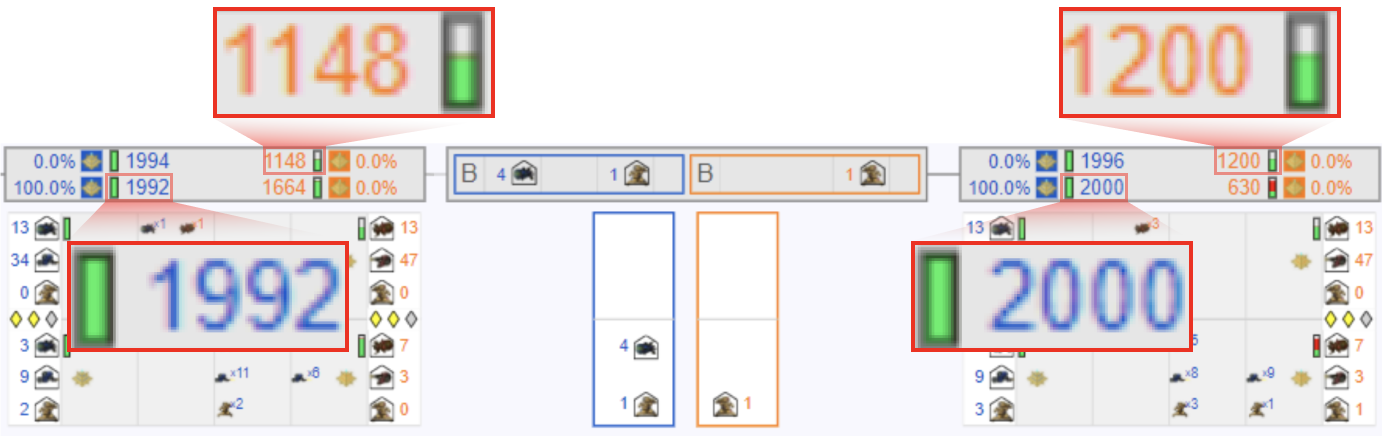}} \label{fig:hp_rises}
\end{tabular}
\\
\cmidrule{2-2}
& 
 \begin{tabular}[t]{@{}p{2.9in} l}
 \textit{Example 2-2.}
 The enemy's marine troop disappears although there is no friendly troop to destroy it. Additionally, the enemy's bottom base loses health even though there are no friendly troops to damage it.
 
 \texttt{\small (outputState.friendlyMarineBldgsBottom +  outputState.friendlyBanelingBldgsBottom}
 &
  \raisebox{-.9\height}{\includegraphics[width=6.5cm]{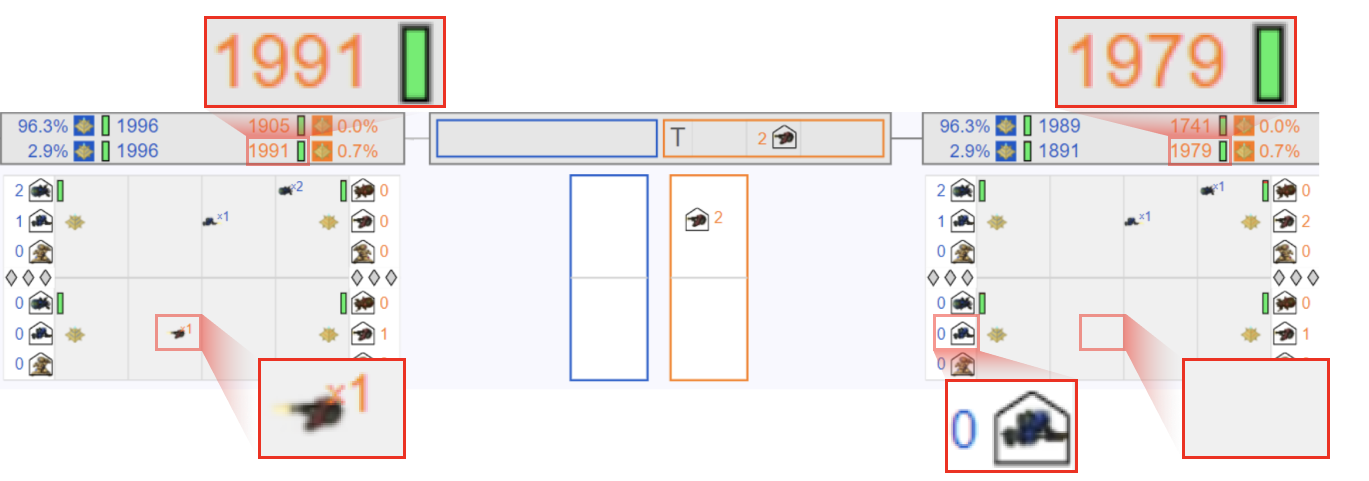}} \label{fig:hp_rises}
  \end{tabular}
\\
&
 \texttt{\small + outputState.friendlyImmortalBldgsBottom = 0) AND}
 
 \texttt{\small outputState.enemyHealthBottom <  inputState.enemyHealthBottom }
 \\
\midrule
\textbf{Symmetry Rules}:

Validate if the agent has learned the symmetric or player-agnostic properties of the domain.
&
\begin{tabular}[t]{@{}p{2.85in} l}
\textit{Example 3-1.}
Reversing the features for the two players where the friendly agent becomes the enemy and the enemy becomes the friendly should produce a perfectly reversed output state. In this example the agent  failed to mirror the immortal troop.

\vspace{3pt}
\texttt{\small outputState.friendlyMarineTopGrid1 > 0 AND outputState.friendlyMarineTopGrid1}
&
\raisebox{-.9\height}{\includegraphics[width=6.5cm]{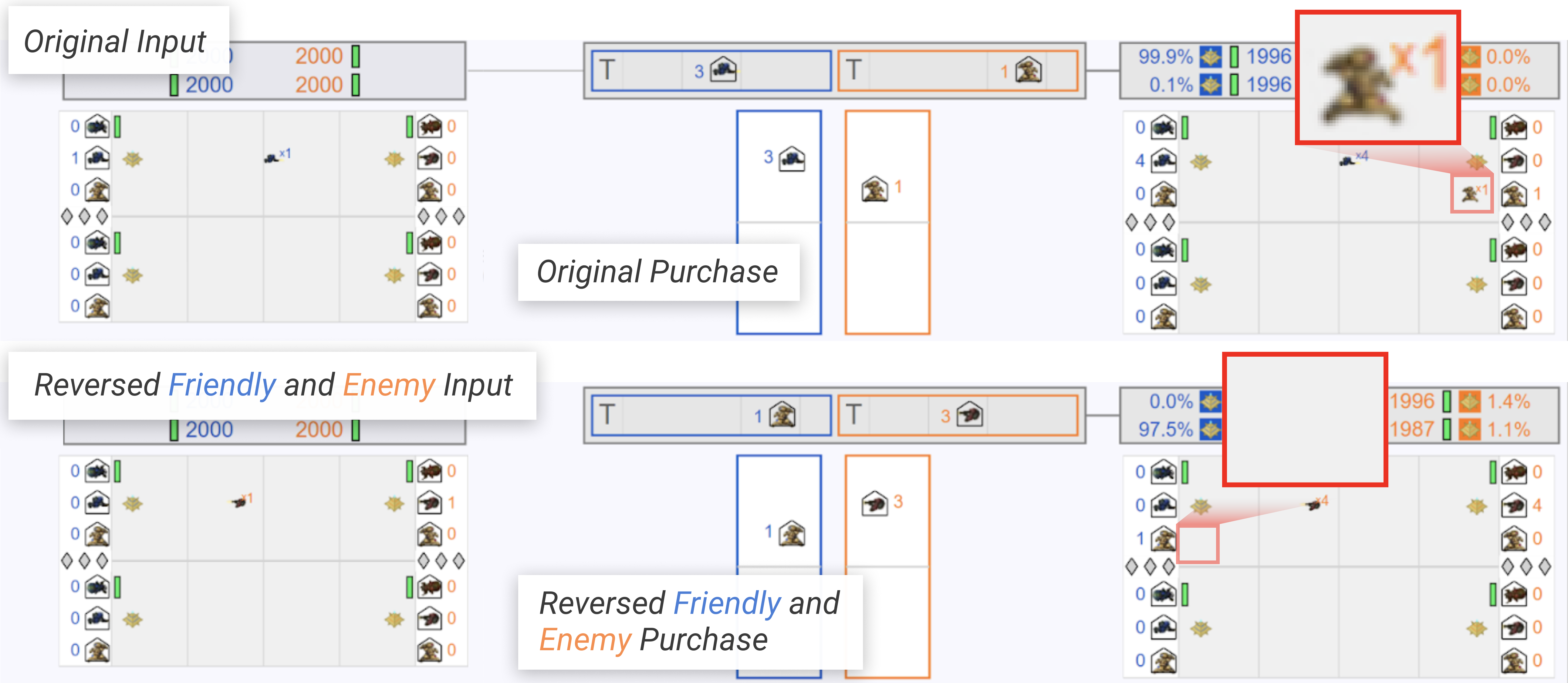}} \label{fig:reverse}
\end{tabular}
\\
&  

\texttt{\small != outputStateForReversedInputs.enemyMarineTopGrid4}
\\
\cmidrule{2-2}
 &
 \begin{tabular}[t]{@{}p{2.85in} l}
 \textit{Example 3-2.}
 Flipping the top and bottom lanes 
 should produce a perfectly flipped output state. In this example, 
 the agent failed to produce a flipped version of the output state (bottom right).
 
 \texttt{\small outputState.friendlyMarineTopGrid2 != }
 
 \texttt{\small outputStateForFlippedInputs. friendlyMarineBottomGrid2 }
 &
 \raisebox{-0.9\height}{\includegraphics[width=6.5cm]{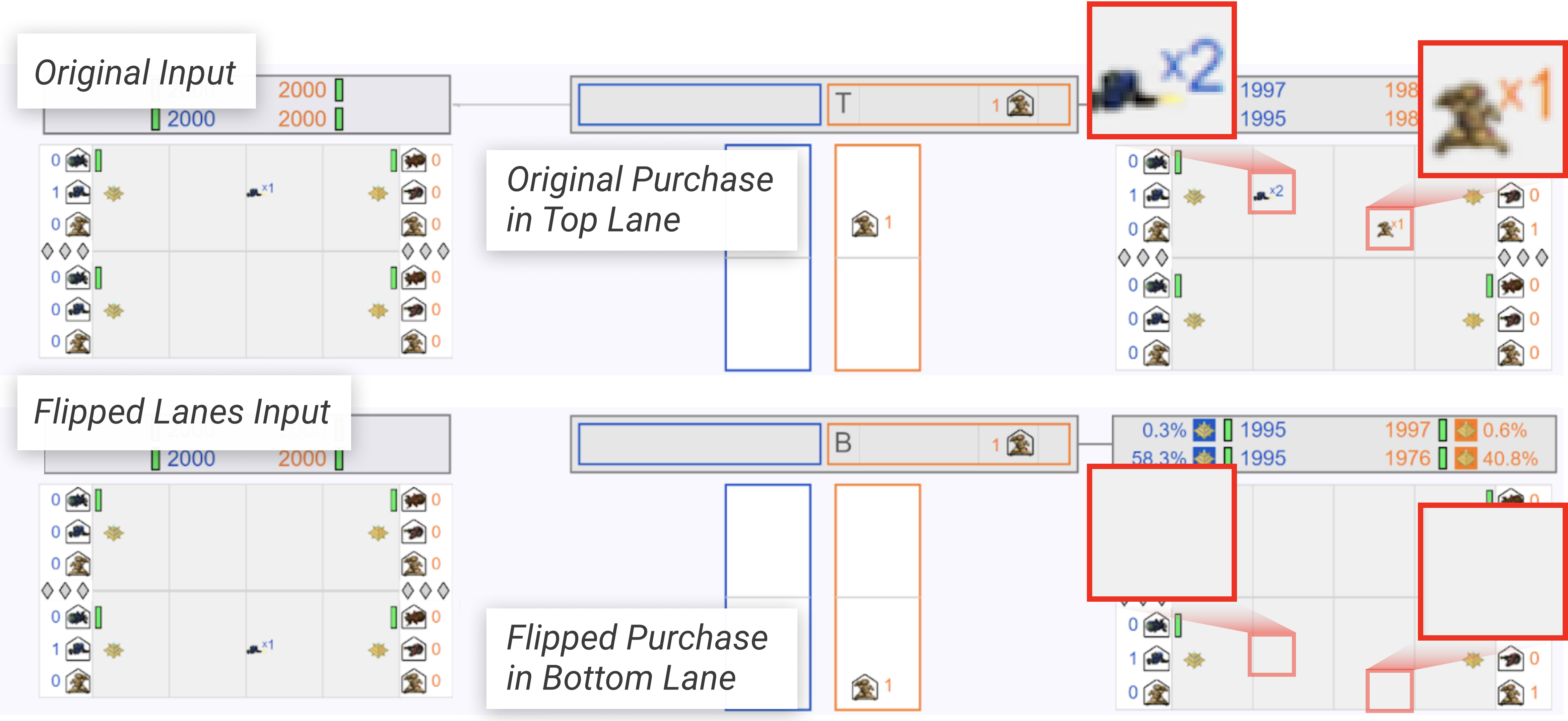}} \label{fig:flip}
 \end{tabular}
\\
\bottomrule
\end{tabular}
\vspace{-5pt} 
\caption{Three classes of C4RL query rules with examples}
\label{table:three-classes}
\end{table*}

\subsection{Adapting \textsc{CheckList} to RL}

While the problems studied in NLP and RL are quite different, we describe how the high-level idea of \textsc{CheckList} for NLP can be adapted to flaw identification for planning-based RL. Conceptually, \textsc{CheckList} for NLP allows a tester to translate common-sense and domain-specific knowledge into sets of examples (i.e. sentences), each with desired system responses that can be checked. Similarly, an RL engineer ``CheckLists'' an RL agent by translating their common-sense and domain-specific knowledge to tests on reasoning steps of an RL agent. Thus, rather than regarding an RL agent's performance by the value of its cumulative rewards on hold-out data, C4RL encourages engineers to also judge an agent based on the quality of its reasoning.

More specifically, \textsc{CheckList} for NLP is based on defining \emph{test types} that are instantiated into \emph{test templates} for example generation. C4RL instead uses \emph{query classes} and \emph{query rules}, which can apply to search trees. While we expect the space of query classes to increase as C4RL evolves, in this work, we support the following three classes.
\label{sec:query-class}
\begin{enumerate}[itemsep=0pt,topsep=1pt]
    \item 
        \textbf{Static State Rules.} Agent search trees contain many abstract states produced by learned knowledge that are also evaluated 
        (e.g. estimating the win probability). Static state rules look for violations of constraints among state variables and value estimates that should always or typically hold. For example, in our ToW domain, there cannot be units of a certain type on the field unless there are production buildings for that unit type. 
    \item 
        \textbf{Transition Rules.} Agent search trees contain state-action-state transitions that should capture the causal dynamics of a domain. Transition rules look for violations of causal properties that should always or typically hold. For example, in ToW, base health can never increase, so transitioning from one state to another should never result in increasing a base's health. 
    \item
        \textbf{Symmetry Rules.} In many planning domains there will be ``common sense" symmetries based on domain knowledge, where correct inferences depend on learning a specific relationship. Similar to metamorphic testing \cite{6963470}, symmetry rules validate the agent has learned such logical consistencies. For example, in ToW, an agent's knowledge and reasoning should be invariant to swapping the top and bottom lanes or switching the sides of the two players. Our C4RL system includes native support for defining such rules. 
\end{enumerate}

To support the construction of rules for an RL domain, we assume a semantically-meaningful domain schema that defines the relevant entities, attributes, and relations. The schema should be rich enough to capture the the states and actions of the agent's search trees. Given a domain schema, query rules can be constructed using a domain-specific language (DSL), which may be application specific or more general purpose. As detailed in Section \ref{sec:interface}, we leverage database technology for this purpose, where query rules are converted to SQL for efficient processing.

Note that by assuming a domain schema, we are restricting the proposed C4RL framework to RL agents that build search trees over interpretable states and actions (e.g. AlphaGo searches over explicit Go positions). This rules out agents that build search trees over learned latent state representations, which are not directly human interpretable (e.g. MuZero \cite{schrittwieser2020}). Such a restriction is currently necessary whenever humans want to interact with an agent's knowledge and reasoning, since the problem of interpreting learned latent representations is in very early stages of research (e.g. \cite{voskuil2021}). Indeed, there are no current approaches that can reliably map such latent representations to human-interpretable schemas over which C4RL query rules could be defined.

\section{C4RL Instantiation for Tug-of-War}
\label{sec:interface}

We describe the C4RL instantiation for ToW used for our evaluation in Section \ref{sec:eval}. While some elements of the instantiation
are necessarily specialized to ToW, the general structure can serve as a schema for C4RL in other domains.

\begin{figure*}[!t]
    \centering
    \includegraphics[width=\textwidth]{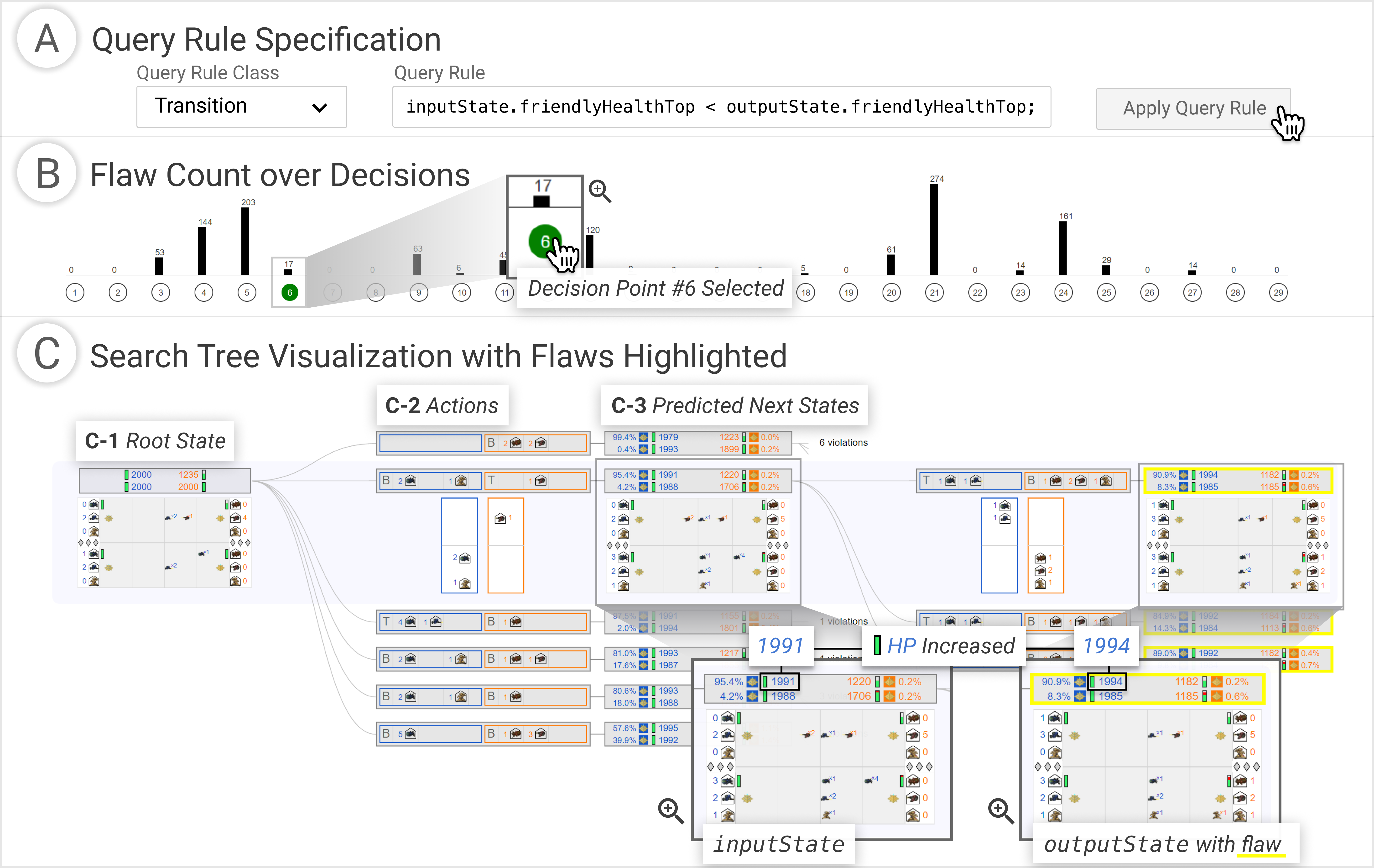}
    \vspace{-10pt}
    \caption{
        C4RL user interface displaying results of a query rule.
        Above, a user constructs a query rule stating the agent's base Health Points (HP) cannot increase (section A).
        After applying this rule, the user selects decision point \#6 which contains 17 flaws (section B).
        This displays a visualization of the matching subset of paths from the full search tree (section C).
        The root state (C-1) depicts the state at the selected decision point. Nodes connected to this state represent Friendly (blue) and Enemy (orange) actions (C-2) and subsequent predicted next states (C-3). Nodes can be expanded to show the full state features.
        Nodes matching this query rule (thus containing a flaw) are highlighted in yellow.
    }
    \label{fig:UI}
\end{figure*}

\subsection{Schema \& Query Language}

Given a collection of game episodes, we store the agent's search trees (one per decision point) into a relational database and transform users' query-rule expressions into SQL queries to retrieve matched records from the databases. In this work, we consider game episodes played against opponents added to the opponent pool during training. 
Our relational schema consists of the following tables: 
\begin{itemize}[itemsep=0pt,topsep=1pt]
{\small
    \item \texttt{\textbf{Episodes} (episodeId, isWin, ...)}
    \item \texttt{\textbf{States} (id, episodeId, decisionIdx, isRoot,  friendlyMarineBldgsTop, friendlyMarineBldgsBottom, ...)}
    \item \texttt{\textbf{Actions} (id, parentStateId, numOfMarineBldgsPurchasedByFriendly, 
    ...)} 
    \item \texttt{\textbf{WinProbabilityOfState} (id, parentStateId, probabilityOfDestroyingEnemyTopBase, ...)}
}
\end{itemize}
This schema allows for representing the agent's tree at each decision point (\texttt{\small decisionIdx}) of each episode (\texttt{\small episodeID}), by encoding each tree path as chains of states and parent actions. The state schema has attributes for describing the agent's state abstraction, including the number of each type of building, current base health, etc. The action attributes encode the action parameters of both the friendly and enemy agent at a decision point, which specify the exact purchases made by the friendly and enemy agent. Importantly, recall that the trees are constructed based on learned transition models. Thus, some state attributes in the trees are produced via predictions from those models. In addition, the agent also makes learned predictions about the probability of each win/loss condition at each state, captured by \texttt{WinProbabilityOfState}. Using this schema, our three rule types, for validating the learned predictions, can be captured as follows 

\textbf{Static State Rules} apply to records in the \texttt{States} table and have a general relational algebra form:
$\sigma_{\textnormal{user-specified rule}} \textnormal{\texttt{\textbf{States}}}$,
where \emph{user-specified rule} is a boolean expression over the numeric state attributes. Table \ref{table:three-classes} gives two examples of such user defined rules. 

\textbf{Transition Rules} apply to predicted state transitions in the search trees, where we refer to the parent state of a predicted transition as the \emph{input state} and the child state as the \emph{output state}. Transition rules have the following general relational algebra form:
$\sigma_{\textnormal{user-specified rule}} \textnormal{\texttt{\textbf{States}}}
\bowtie
\textnormal{\texttt{\textbf{Actions}}}
\bowtie
\textnormal{\texttt{\textbf{States}}}$
where the first \texttt{States} represents the input state, the second \texttt{States} represents the output state, and $\bowtie$ is the \textit{join} operator. Again, \emph{user-specified rule} is a boolean expression over the numeric attributes of the input/output states and action. Table \ref{table:three-classes} gives two examples.

\textbf{Symmetry Rules} differ from the above classes in that they involve testing of the agent's prediction on counterfactual states that were not in the original episodes. An example is to check if the win probabilities of the agent change significantly if the top and bottom lanes were swapped. While there exist many ways to support this, we create additional database tables that store attributes of counterfactual states and actions resulting from symmetry transformations of the original states and actions.
Specifically, we perturb all static states and their corresponding action pairs, producing flipped or reversed versions of the originals. We then run inference on all flipped or reversed static states and actions, and include a reference in the schema linking the inference result of a flipped or reversed state and action to its original result. 
The relational algebra form of the queries is like that of transitions, except the join involves an original state-action-state and corresponding transformed state-action-state. The \emph{user-specified rule} can then be any boolean expression that compares information between the original and transformed transition data. Table \ref{table:three-classes} gives two examples.

\subsection{Visualization of Search Trees}

To help users examine the results from C4RL queries, we developed a visualization of the agent's search trees by adapting our earlier work~\cite{khanna2022finding,tabatabai2021did}.
A tree starts with the root node (Figure \ref{fig:UI} C-1), which represents the state at the selected decision point.
Each state node has a compact view (smaller box with 4 health points for each base) and a larger state view (graphical representation with thumbnail of the game map) that can be expanded by the user.
Nodes connected to the root node in the tree show combinations of friendly (blue) and enemy (orange) actions (at C-2) and states the friendly agent predicts next (at C-3).
Figure~\ref{fig:UI}-C shows 6 children of the root node, where  one of them is expanded, while the remaining five are in the compact view.  
For the expanded example (right of C-1), the friendly agent's action is buying 2 baneling buildings and 1 immortal building in the bottom lane and the enemy agent's action is buying 1 marine building in the top lane. 
The agent's prediction of the state at the next decision point (30 seconds after action) is displayed to the right of the corresponding action (C-3). Each predicted state includes the predicted win probabilities at the top of the state. The four percentages give the probability that the friendly/enemy agent will win in the top/bottom lanes. For example, at C-3 the predicted state shows the agent is 95.4\% confident it will end up winning the game from the top lane.
This state node is followed by another pair of actions, and the predicted grandchild state (leaf of the search tree). 

\subsection{C4RL Interface}

We design and build an interactive C4RL user interface for specifying query rules and visualizing violations.
For purposes of the user study, the current interface is optimized to explore a single selected game at a time, though it is straightforward to support multiple game repositories. 
The interface (shown in Figure~\ref{fig:UI}) consists of three sections: \textbf{A.} Query Rule Specification, \textbf{B.} Flaw Count over Decisions, and \textbf{C.} Search Tree Visualization with Flaws Highlighted.
Figure~\ref{fig:UI} illustrates a case of checking whether the agent's transition predictions violate the \textit{monotonicity} constraint that the Health Points (HP) of a base can never increase. 

\textbf{A. Query Rule Specification.}
The user can specify the monotonicity-checking rule using the first section (at Figure~\ref{fig:UI}-A). 
They specify the rule class using the dropdown menu, (i.e., Transition)
and the rule using a SQL-like statement: 
{\small \texttt{inputState.friendlyHealthTop < outputState.friendlyHealthTop}}.
The interface provides a visual guide for users to simply click components to auto-complete their attribute names.

\textbf{B. Violations over Decisions.}
Upon clicking the apply button, the system internally transforms the rule into an SQL query and runs it over the database.
A bar chart is shown to denote the number of query-rule violations in the tree at each decision point during the game
(depicted at Figure~\ref{fig:UI}-B). 
The user can then select decision points for a deeper investigation of violations in a particular search tree. 

\textbf{C. Search Tree Examinations.}
Upon selecting a decision point,
the interface displays a subset of the search tree where query-rule matching nodes are highlighted in yellow. In Figure~\ref{fig:UI}-C, grandchildren nodes on the right are highlighted because the base's health (HP = 1994) is greater than that of the parent state (HP = 1991) which match the query rule specified by the user. In other words, this is a flaw where health monotonicity is violated.

Users may perform further analysis by expanding the collapsed representation of the nodes (or the sub-trees), to examine the details of the other highlighted violations or other decision points.
The end result of this exploration might be a report that gives concrete examples of violations, which AI engineers can use to judge the severity of violations, locate the cause of faults, and plan for potential fixes (e.g. additional training in specific scenarios)\cite{DBLP:journals/corr/abs-1809-07424, khanna2022finding}. 

\section{Qualitative Experimental Study}
\label{sec:eval}

We present two qualitative investigations for C4RL in the context of Tug-of-War. We describe a case study where agent developers used C4RL to discover agent flaws and
a user study we conducted with 7 human-subject participants.

\subsection{Agent Developer Case Study}

A subset of the authors who developed the agent ``CheckList'' it using C4RL.
For a learned agent which is able to win 80\% to 90\% of ToW games against different model-free agents, we selected one game replay where the agent lost.
This game consists of 29 decision points and 103,520 transition inferences, where a single search tree has a maximum of 1,227 transitions, of which contain 3,680 static states.
Using C4RL, developers investigated how the agent's reasoning flaws at the inference level contributed to its defeat.

\textbf{No buildings but units on the field (Example 1-1).}
Example 1-1 in Table \ref{table:three-classes} depicts an abstract state produced by the agent where it expects to have an immortal unit on the field; an impossible inference because the agent has not built any buildings to produce this unit.
This query rule finds a total of 663 unique instances of this static state flaw in the entire game from 3 of 29 decision points.
There are 611 static state flaws occured at decision point 29 (the last decision point in the game before the agent loses) out of 3,680 static states. 
This type of error could completely change the evaluation of a static state and its upstream transitions. 
The developers hypothesize the agent has not experienced enough situations with immortals as the agent rarely uses immortals or 
has not learned the unit/building association.
Modifying this query rule to check whether this flaw occurs with another troop, banelings,
yields only two static state violations in the entire game, all occurring at decision point 2.
This shows the agent is missing some common-sense regarding production of baneling troops too, but the static state violations involving the immortal troops are more prevalent.
This could be due to the fact that the agent has more training experience with banelings. Banelings are cheaper to purchase and are bought more frequently than immortals.

\textbf{Monotonicity violation (Example 2-1).}
\label{sec:caseStudy2-1}
In Tug-of-War, the health of a base can never increase. Example 2-1 in Table \ref{table:three-classes} checks for non-trivial violations of this monotonicity property 
for the enemy's top base. It found 26,540 violations out of all 103,520 predicted transitions in the game trees. The violations occur at almost all decision points, except for the first two. 
Interestingly, in the specific flaw instance shown in Example 2-1, the agent's bottom base also violates the monotonicity property. 
Modifying this prompt to instead check monotonicity for the agent's top base finds 2,403 violations dispersed across 5 decision points.   
This is a serious error that clearly demonstrates the the agent has not fully learned some key constraints of the game. 
The developers noticed that the flaws occur most frequently in states where the agent's base has lower health, which suggests more agent training is needed in such situations.

\textbf{Phantom Health Decrease (Example 2.2).}
If an agent does not have any buildings, the opponent's base should have full health (2000), because no troops are present to damage it. This fact is expressed by the query rule shown in Example 2-2 in Table \ref{table:three-classes} which returns 6,311 violations, all occurring in the first five decision points. This is  another indicator that the agent has not fully captured common sense constraints about the health dynamics of bases. 

\subsection{User Study}
\label{sec:user-study}

We conducted an in-person qualitative user study to investigate how researchers and engineers with AI expertise, but no prior experience with the agent or domain, would use C4RL.

\subsubsection{Participants and Protocol}
Recruited participants included 7 graduate students at our university (2 M.S. and 5 Ph.D), who had taken a course on Reinforcement Learning and did not have previous experience with the task, domain, or the C4RL interface.

Participants were invited to a two hour session in a controlled laboratory environment and were compensated with a \$20 gift card at the conclusion of the study.
All sessions began with a 30 minute tutorial covering the ToW domain, agent architecture basics, and one query-rule example for each query-rule class. 
Participants were then asked to use the C4RL interface to construct as many query-rules as they could for the remainder of the session.
To help formulate query-rules, participants were provided a cheat-sheet with examples from the tutorial.
Participants recorded their query-rules and described the intent of their rules in a spreadsheet.

\subsubsection{Analysis} User-study result analysis was conducted based on the following five research questions.

\textbf{Q1. Are participants able to produce correct rules?}
We consider three cases of rule correctness. \emph{Sound rules} are logical consequences of the domain constraints. Violation of a \emph{sound rule} imply said violation is true a reasoning flaw. \emph{Suspicion rules} correspond to common-sense game conditions that will usually hold, thus a violation should be investigated. Finally, \emph{unsound rules} are invalid rules that indicate the participant formed an incorrect assumption; violations of aforementioned do not reflect agent flaws.

Participants constructed a total of 126 query-rules. The research team analyzed their correctness and found 16 unsound, 98 sound, and 12 suspicion rules. This indicates participants, given a short tutorial, were able to ``checklist'' the agent.

\textbf{Q2. What types of rules did participants specify?}
Researchers grouped the 110 sound or suspicion rules into multiple categories, through \textit{affinity diagramming}, a well-known technique used in user study analysis.
Two of the authors went through the 110 rules independently, assigned categories based on the type of RL violations they would detect,  
and discussed standard category names to resolve disagreed rule categorizations. 
The following four high-level rule categories were identified. 
\begin{itemize}[itemsep=0pt,topsep=1pt]
  \item \textbf{Monotonicity Violation.}
  While the values of certain features must behave monotonically, the agent's output (e.g. predicted state) may not follow this property.
  For example, base health, a feature for a state, cannot increase.

  \item \textbf{Value Out Of Range Violation.}
  The values of some features must be within certain ranges defined by domain constraints.
  For example, base health must always be within a range of 0 to 2000.

  \item \textbf{Causal Violation.}
  The values of some features can only be changed when other features are changed.
  An example of causal violation includes when the agent produces a state where there are troops on the field, but there is no buildings that can produce them.

  \item \textbf{Symmetry Violation.}
  This includes all cases for the \textit{symmetry rules} class.
  When inputs are flipped, nearly perfectly flipped outputs are expected.

\end{itemize}
Table \ref{table:Violations-Categories} further decomposes the high-level categories into sub-categories that are Tug-of-War specific and provides the count for each. The most common rule type was \textit{causal violations} (61 rules), followed by \textit{symmetries} (22 rules), \textit{value-out-of-range} (20 rules), and \textit{monotonicity} (7 rules).
There were 6 sub-categories of causal violations, some of which were surprising. 
Notably, participants identified a number of rules that captured  
situations when the agent erroneously predicted a win/loss (i.e., ``\textit{not supposed to win/lose}'').
For example, participant \#1 specified a causal relationship between the chance of winning and the agent's base health as ``\textit{[If the base health of friendly AI in the top lane is equal to zero, then the chance of winning for friendly AI in the top lane should be zero.]}'' The fact that the agent violated such a rule was surprising and clearly a point of concern.

\begin{table}[!tb]
\centering
\begin{tabular}{p{0.75in}|p{1.8in}|@{}c@{}}
\toprule
\textbf{Category}
& \textbf{Sub-categories} {\small(domain-specific)}
& \textbf{Count}
\\ \hline

\multirow{2}{*}{Monotonicity} 
 & Base health (increasing) & 4   \\ 
  \cline{2-3}
 & Building count 
 &3\\
  \hline
 
 \multirow{3}{*}{\shortstack[l]{Value Out\\of Range}} 
 & Base health & 12\\ 
 \cline{2-3}
 & Win probability &  7\\ 
 \cline{2-3}
 & Number of units &1
 \\ 
 \hline
 
  \multirow{5}{*}{\shortstack[l]{Causal\\Violation}}
 & Not supposed to win/lose  &26\\ \cline{2-3} 
 & No building but troops   &19 \\ \cline{2-3} 
 & No building but health change  &7 \\ \cline{2-3} 
 & Unit positioning  &5 \\
 \cline{2-3} 
 & No troops but health change  &3    
 \\ \cline{2-3} 
 & Rock-paper-scissors &1 \\ 
 \hline
 
  \multirow{4}{*}{Symmetries} 
 &  Base health 
 &9     \\ \cline{2-3} 
 & Win probability  &8 \\ \cline{2-3} 
  &Unit count &4  \\ \cline{2-3} 
 &  Building count  &1\\ 
 \bottomrule
\end{tabular}
\vspace{-5pt}
\caption{
Violation Categories Table: 
Four categories of high-level RL violations and 15 sub-categories according to the domain. The rightmost column displays total rule counts constructed by participants.
}
\label{table:Violations-Categories}
\end{table}

\textbf{Q3. Can participants  creatively construct new rules?}
Participants were given two types of guidance: a handout of example rules and screenshots of trees with flaws.
We analyzed how many of the participant rules were variations of the guidance versus original constructions. Each rule was judged whether it was ideated from: 1) handout, 2) a screenshot, or 3) none of these (i.e. original). 
Of the 110 sound or suspicion rules, only 5 were direct adaptations of rules explicitly given in the handout and 30 were constructed based on flaws in the tree examples provided. The remaining 76 rules (68\%) appeared to be original and functionally distinct from the provided examples. This included a number of rule forms that the research team had not previously considered.
For example, participant \#1 reported a suspicion rule regarding the ``Rock-Paper-Scissors'' relationship among the three types of troops as ``\textit{If the friendly AI has some Banelings and the enemy AI did not purchase any Immortal buildings, it should be guaranteed that the enemy base health is going to be reduced.}''
Participants also expressed rules in English that the current version of the C4RL syntax does not support. For example, participant \#3 wrote, ``... a violation type to combine player swap and lane swap. ... When both players and lanes are swapped, friendly top base health become enemy bottom base health.''

\textbf{Q4. How diverse are the rules formed by each participant?}
Overall there were a diversity of rules among the 4 categories and 15 sub-categories. However, this leaves the question of the diversity of rules for individual participants. 
To address this, 
we counted the number of unique sub-categories generated by each participant. 
On average participants form rules from 6.7 categories and all participants constructed at least four unique sub-categories of rules. This shows that participants were naturally considering different types of relationships throughout their investigation. 
We observed that the participants differed in their quantity and diversity of query rules and the strategies to find rules.
For example, participant \#2 constructed 41 sound or suspicion rules, the highest quantity among all participants.
They focused exhaustively on checking all bounds for a narrow set of features (e.g., out of range violations).
On the other hand, although participant \#6 
constructed only 13 rules (which is less than the average), they successfully created rules from 10 different sub-categories. 
This suggests further study into the types of strategies taken by C4RL users and how they might be guided toward the most effective strategies.  

\textbf{Q5. Are there any patterns in the participants' exploration?}
Lastly, we wondered if there exist any interesting patterns in the list of query-rules which participants constructed. 
One interesting observation is that
some participants
formed query-rules based on what they previously constructed.
For example,
after participant \#2 formed a rule as ``\textit{If friendly AI dies in the top lane, then enemy wins,}''
they constructed 7 query-rules that have almost the same structure, such as
``\textit{If friendly AI dies in the \textbf{bottom} lane, then enemy wins},''
``\textit{If \textbf{enemy} dies in the top lane, then \textbf{friendly AI} wins},'' and so on.
This implies that some of the rules have many variations that can easily be populated systematically. 
An interesting future direction would be to create templates for the rules, so that the system automatically creates or suggests combinations for rules so that the users do not need to construct variations of the same rule manually.

\section{Summary}
We present CheckList for Reinforcement Learning (C4RL), a behavioral testing methodology adapted for RL agents; allowing researchers to evaluate an agent beyond some measurement of expected value.
We formalized three classes of query-rules and developed a C4RL system in the context of a planning-based agent for a real-time strategy game. 
Our qualitative studies demonstrates users can use C4RL to identify a diverse range of flaws in the agent's reasoning.
Overall, C4RL is a straightforward yet effective approach for validating RL agents, and there are several possible variations to be explored.

\section*{Acknowledgements}
This work was supported in part by DARPA (N66001-17-2-4030), Google (GCP19980904), and NAVER AI Lab.

\newpage

\bibliography{paper.bib}

\end{document}


\maketitle

\section{Agent Learning Approach Details}


To learn the required components we use model-free RL to learn a Q-function $Q(s,a)$ that estimates the value of action $a$ in abstract state $s$. 
Specifically, since this is a two-player game, we use a tournament-style self-play strategy, where a pool of previously trained model-free agents, each with their own Q-function, is used to play against a currently learning agent. The agent is trained until it achieves a high win-percentage against the pool or training resources are expended. This is similar to the pool-based self-play strategy employed by AlphaStar for the full game of StarCraft 2 \cite{vinyals2019}. 

To train each agent we use a variant of DQN \cite{mnih2015} called Decomposed Reward DQN \cite{juozapaitis2019}, allowing us to learn a Q-function that returns a vector of probabilities over the different endgame possibilities (e.g. winning by destroying the opponents top base). The sum of the win-condition probabilities for a specific player is the overall value of the action for that player (i.e. the win probability). This vector provides more insight into the agent's decision making and part of the visualization in our explanation interface. The Q-function of the best agent in the pool (typically the last trained agent) is used as the learned action ranking for search. In addition, it is also used for the state-evaluation function $V(s)$ by letting the state value to be the value of the best action, that is $V(s) = \arg\max_a Q(s,a)$.

We represent the Q-function using a 3 layer feed-forward neural network with an input consisting of features describing the abstract state and action. The network outputs the predicted value vector of the state-action pair. Self-play training was conducted for two days, after which the learned model-free agent appeared to be quite strong, likely better than most humans with some game experience.


To learn the dynamics model used for search we formed a training set of abstract state transitions observed in games between agents during pool-based training in addition to games involving random agents to further increase data diversity. Each data instance was of the form $(s, a_f, a_e, s', \vec{r})$ giving the current abstract state, friendly action, enemy action, next abstract state, and decomposed-reward vector respectively. Here the reward vector is the zero vector for all states, except at the end of the game where it is the zero vector for a loss and a one-hot encoding of the win condition otherwise. We designed a feed-forward neural network that takes $s$, $a_f$, and $a_e$ as input to predict $s'$ and $\vec{r}$. Note that this approximates the dynamics as a deterministic function. While the actual dynamics are stochastic due to unit level randomization of damage, in aggregate, a deterministic model appears to be adequate for strong play. 

The final planning-based agent using the learned components is able to win 80\% to 90\% of games against the different model-free agents added to the pool during training. This shows that despite potential inaccuracies in the learned components, look-ahead search is able to provide significant improvement over the model-free agents.

\bibliography{paper.bib}